\documentclass[sn-apa]{sn-jnl}

\DeclareMathOperator*{\argmin}{arg\,min}
\usepackage{natbib}

\jyear{2021}%

\theoremstyle{thmstyleone}%
\theoremstyle{thmstyletwo}%

\theoremstyle{thmstylethree}%

\begin{document}

\title[Multi-view graph structure learning using subspace merging on Grassmann manifold]{Multi-view graph structure learning using subspace merging on Grassmann manifold}


\author[1]{\fnm{Razieh} \sur{Ghiasi}}\email{raziehghiasi@gmail.com}

\author*[1]{\fnm{Hossein} \sur{Amirkhani}}\email{amirkhani@qom.ac.ir}

\author[2]{\fnm{Alireza} \sur{Bosaghzadeh}}\email{a.bosaghzadeh@sru.ac.ir}

\affil[1]{\orgdiv{Computer and Information Technology Department}, \orgname{University of Qom}, \orgaddress{\city{Qom}, \country{Iran}}}

\affil[2]{\orgdiv{Artificial Intelligence Department }, \orgname{Shahid Rajaee Teacher Training University}, \orgaddress{\city{Tehran}, \country{Iran}}}


\abstract{Many successful learning algorithms have been recently developed to represent graph-structured data. For example, Graph Neural Networks (GNNs) have achieved considerable successes in various tasks such as node classification, graph classification, and link prediction. However, these methods are highly dependent on the quality of the input graph structure. One used approach to alleviate this problem is to learn the graph structure instead of relying on a manually designed graph. 
In this paper, we introduce a new graph structure learning approach using multi-view learning, named MV-GSL (Multi-View Graph Structure Learning), in which we aggregate different graph structure learning methods using subspace merging on Grassmann manifold to improve the quality of the learned graph structures.
Extensive experiments are performed to evaluate the effectiveness of the proposed method on two benchmark datasets, Cora and Citeseer. Our experiments show that the proposed method has promising performance compared to single and other combined graph structure learning methods. }




\maketitle

\section{Introduction}\label{introduction}
Graphs are known as a powerful tool for representing complex relationships between non-Euclidean data
such as social networks, biological networks, e-commerce networks, communication networks, and sensor networks. Conventional deep learning algorithms cannot be directly applied to graph data because this type of data is inherently complex and irregular. Thus, a new class of methods called Graph Neural Networks (GNNs) has been developed to apply deep learning methods to graph data \citep{keramatfar2022graph}. 

GNNs use an information diffusion mechanism to refine the nodes' representation. 
According to this mechanism, each node aggregates the information of its neighbors and combines the aggregated data with its feature vector. This process is repeated $k$ times to obtain a new representation of each node according to the structural information of its $k$-hop neighbors \citep{liu2020introduction, wu2020comprehensive, zhang2020deep,zhou2020graph}. These methods utilize both node feature and network topology in an interactive learning process. 
Based on the used aggregation process, there are several variants of GNNs, such as Graph Convolutional Networks (GCN) \citep{bruna2014spectral,kipf2017semi,defferrard2016convolutional}, Graph Attention Networks (GAT) \citep {velickovic2018graph, zhang18}, and Graph Autoencoder Networks (GAE) \citep {kipf2016variational}.

Recent studies show that the performance of GNNs is highly dependent on the quality of the used graph structure \citep{zhu2019robust,jin2020graph,fox2020robust,jin2021adversarial,zugner2018adversarial, shanthamallu2020regularized, dai2018adversarial, zhan2020graph}. 
In many applications, a sound and complete graph is not available. 
In some other applications, such as in natural language processing tasks, the data is not inherently graph-based. 
In these cases, the graph structure should be learned from the observed data. The simplest way to create a graph structure is to calculate dependencies between samples using a similarity function, and exploit a thresholding method, such as K-Nearest Neighbors (KNN). However, these methods face several important challenges, such as choosing an appropriate similarity function and selecting the value of the parameters (e.g., $K$), which can have a profound effect on the quality of the learned graph. 

In recent years, many methods have been proposed for Graph Structure Learning (GSL) such as GLCN \citep{jiang2019semi}, ProGCN \citep{jin2020graph}, and GRCN \citep{yu2020graph}, to obtain more quality graphs. Most existing methods learn a single relationship between pairs of nodes. However, in many applications, there are several types of relationships between two nodes, such as the multi-relational structure of a knowledge graph; different protein-protein interactions in biological systems; and friendship, acquaintance, and business relations in social networks. Therefore, learning only a single graph cannot completely show the relationships between nodes. 

Existing GSL methods utilize different techniques to discover similarities between nodes. As a result, each method explores a different aspect of the data. 
Therefore, when different methods are integrated, they can provide more comprehensive
knowledge of the data, which can result in improving the generalization capability of the model. In addition, the usage of multiple views can alleviate the negative impact of noise \citep{wan2022self, yan2021deep}. This paper introduces Multi-view Graph Structure Learning (MV-GSL) method to merge different learned graphs based on Grassmann manifold \citep{dong2013clustering} to exploit the multi-view knowledge. 

In summary, this work has the following contributions:

\begin{itemize}
\item We conduct an extensive study on single-graph structure learning methods. 
\item We propose a new graph structure learning approach that integrates multiple graphs using subspace merging. The adopted graphs are automatically constructed using the existing single-graph structure learning methods. The aggregated graph is used in a node classification task. 
\item Comprehensive experiments on two benchmark datasets show that our proposed method achieves better or highly competitive results compared to the competitors.
\end{itemize}
The rest of this paper is structured as follows. Section \ref{backgraound} describes the theoretical basic concepts. Related work is reviewed in Section \ref{Relatedworks}. The key idea of the proposed method is introduced in Section \ref{ProposedFramework}. Section~\ref{Experiments} presents experimental results and analysis. Finally, the paper is concluded in the Section~\ref{Conclusion}. 

\section{Background}\label{backgraound}
In this section, we introduce the related theoretical basis, including the graph convolutional network, Graph structure learning, and multi-view learning.

\subsection{Graph}

In this paper, a graph is represented as \(G = (V,E)\), where \(V \in R^n\) is a set of nodes and \( E \subseteq \{(i,j) \vert i,j \in V\}\) is a set of relational edges. Each node $i$ has a set of features or a signal \(x_i\in R^d\) and a label  \(y_i \in R^c\). In addition, each edge can have a set of features \( e_i\in R^f\). Here, $n$, $c$, $d$, $f$ are the number of nodes, the number of classes, the size of feature vector of nodes, and the size of feature vector of edges, respectively. The structure of a graph is determined by the adjacency matrix \(A \in R^{n \times n}\), where if there is an edge between two nodes, the corresponding element is greater than 0, and otherwise it is equal to 0. In addition to the adjacency matrix, the graph structure can be represented using the Laplacian matrix \(L\in R^{n \times n }\), which is defined as follows:

\begin{equation}
\label{eq:eq1}
\begin{aligned}
L=D-A, 
\end{aligned}
\end{equation}
Where \(D\in R^{n \times n}\) is a degree matrix whose diagonal elements are equal to the degree of each node and non-diagonal elements are 0. To create numerical stability in deep models, the symmetric normalized version of Laplacian matrix is usually used. It is defined as follows:
\begin{equation}
\label{eq:eq2}
\begin{aligned}
L=D^{\frac{-1}{2}} LD^{\frac{-1}{2}}=I-D^{\frac{-1}{2}} AD^{\frac{-1}{2}}.
\end{aligned}
\end{equation}

The Laplacian matrix has different valuable and important properties. For example, when the adjacent matrix is non-negative and symmetric, its Laplacian matrix is a real symetric positive semi-definite matrix and has a full set of orthonormal eigenvectors. Therefore, it can be rewritten as follows:
\begin{equation}
\label{eq:eq3}
\begin{aligned}
L=U\Lambda U^T,
\end{aligned}
\end{equation}
where \(U \in R^{n \times n}\) is the matrix of eigenvectors and \(\Lambda\) is the diagonal matrix of eigenvalues. The Laplacian matrix can also generalize the concept of Fourier transform into graph signals through spectral decomposition \citep{dong2016learning}. 

\subsection{Graph Neural Network}
\begin{figure}
\centering
\includegraphics[width=0.9\textwidth]{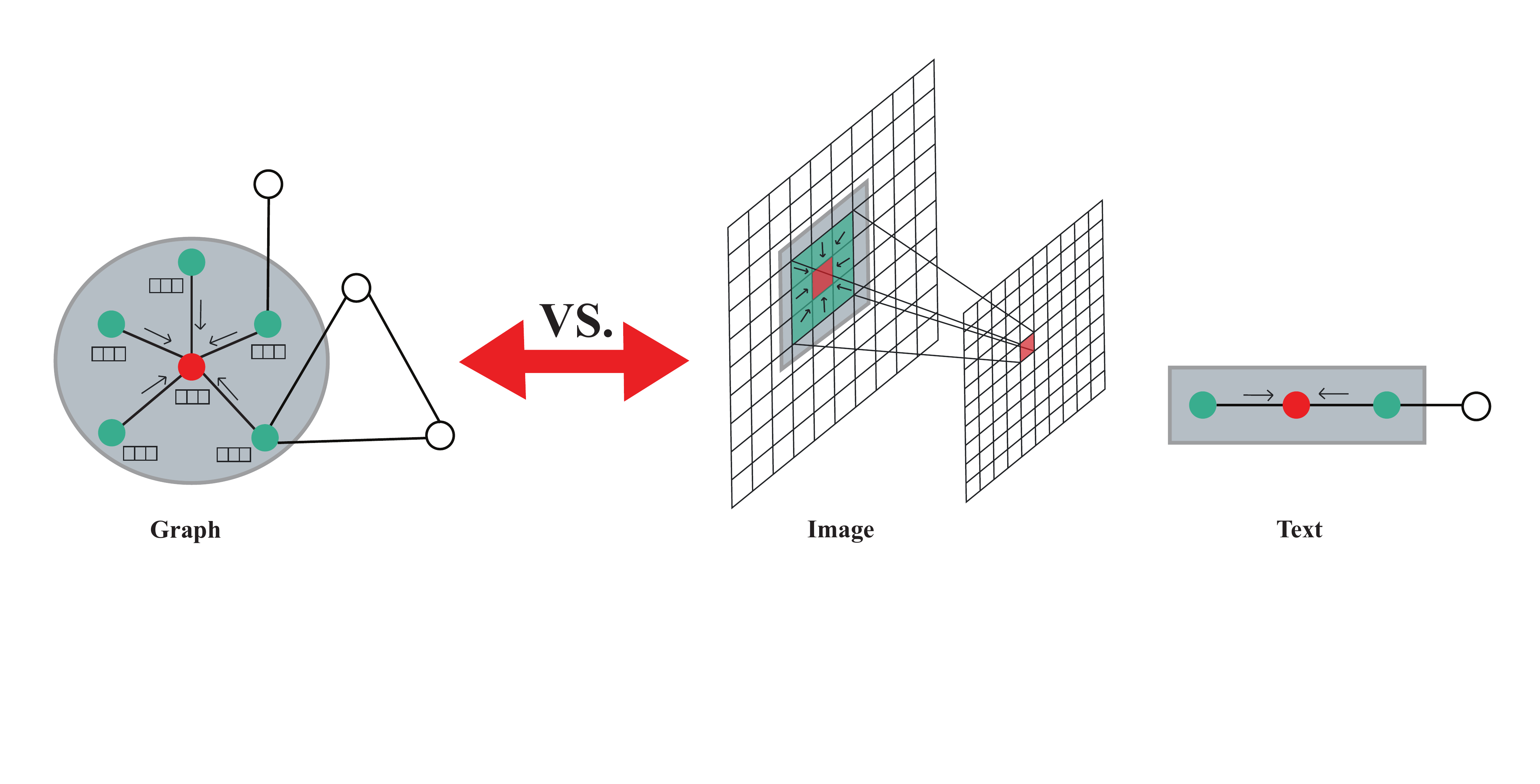}
\caption{\label{fig:GCN} Convolution operation in Euclidean data (image or text) (right) and non-Euclidean data (left)}
\end{figure}

Graph Convolutional Networks (GCNs) are one of the most popular types of GNNs. They generalize existing convolution operations for the Euclidean data (e.g., image, text, etc) to the non-Euclidean data \citep{wu2020comprehensive}. Convolution operations in the Euclidean data learn new properties from adjacent pixels/words, similarly, convolution operations in graphs seek to learn new features from adjacent nodes (See Figure \ref{fig:GCN}).

In literature, two strategies have been used to define convolution filters. These strategies create two categories of graph convolution networks, including \textit {spectral or frequency-based methods} and \textit {spatial-based methods}.

\subsubsection{Spectral Methods}
The purpose of spectral GCN is to define graph convolution using Fourier transform. In the Fourier domain, the convolution operation is calculated by eigendecomposition of the Laplacian matrix \citep{liu2020introduction} as follows:
\begin{equation}
\label{eq:eq4}
\begin{aligned}
h=x\ast g=\mathcal{F} ^ {-1} ( \mathcal{F} (x) \bigodot \mathcal{F} (g))=U(U^T g \bigodot U^T x),
\end{aligned}
\end{equation}
where \(x\) is the input feature vector (signal graph) and $h$ is the updated feature vector, \(\mathcal {F}\) indicates Fourier function, and \(U\) is eigenvectors matrix of the normalized Laplacian matrix \( (L=I_n-D^{\frac{-1}{2}} AD^{\frac{1}{2}}=U \Lambda U^T) \). By defining the spectral filter as \(g_w= diag (U^T g)\), the graph convolution can be simplified as follows:
\begin{equation}
\label{eq:eq5}
\begin{aligned}
h=x*g_w=Ug_w U^T x.
\end{aligned}
\end{equation}

Therefore, different types of spectral convolution networks can be formed depending on the filter \(g_w\). For example, \cite{bruna2014spectral} proposed the spectral CNN, which considers the filter as a diagonal matrix of learnable parameters \(g_w=w_{ij}\). Therefore, the graph convolution operation is defined as follows:
\begin{equation}
\label{eq:eq6}
\begin{aligned}
h_{:,j}=x_{:,j} \ast g_w=\sigma (\sum_{i=1}^{d_{l-1}} Uw_{i,j}
U^T x_{:,j}), \qquad \qquad j=1,2,…d_l,
\end{aligned}
\end{equation}

In this equation, \(d_l\) is the number of output channels and \(\sigma\) represent a non-linear activation function. However, this operation leads to high computational complexity, non-scalability for large graphs (due to the need to calculate the eigenvectors of the Laplacian matrix), and non-local filters (due to the use of non-parametric filters). To solve these challenges, \cite{defferrard2016convolutional} proposed ChebNet, which approximates the convolution filter using \(K^{th}\) order Chebyshev polynomials, i.e., \(g_w=\sum_{k=0}^K w_k T_k (\hat{\Lambda})\) where \( \hat{\Lambda} =\frac{2 \Lambda} {\lambda_max} -I_n\) and \(\lambda_{max}\) is the largest eigenvalue. Then, the convolution operation is calculated as follows:  
\begin{equation}
\label{eq:eq7}
\begin{aligned}
h=U(\sum_{i=0}^k w_i T_i (\hat{\Lambda}))U^T x= \sum_{i=1}^k w_i T_i (\hat{L})x,
\end{aligned}
\end{equation}
where \(L= \frac{2L} {\lambda_{max}} - I_n\)  and \(T_k (x)=2xT_{k-1} (x)- T_{k-2} (x)\), with \(T_0 (x)=1\), \(T_1 (x)=x\). 
This method solves the localization issue and reduces computational complexity. However, it can lead to the overfitting issue in local neighborhood structures for graphs with high degree distributions because it does not limit the polynomial order (\(k\)). To solve this problem, \cite{kipf2017semi} have presented a simpler version of ChebNet, which limites the order of Chebyshev polynomials to 1 (\(k = 1\)) and approximates the largest eigenvalue (\( \lambda_{max}= 2 \)). In the graph network literature, this network is known  as the GCN. In this method, the convolution operation is simplified as follows:
\begin{equation}
\label{eq:eq8}
\begin{aligned}
h=\sum_{i=0}^1 w_i T_i (\hat{L})x =w_0 T_0 (\hat{L})x + w_1 T_1 (\hat{L})x=\sigma (w(I_n+\hat{D}^{\frac{-1}{2}} \hat{A} \hat{D}^ {\frac{-1}{2}} )x),
\end{aligned}
\end{equation}
where \(\hat{A}=A+I\) and subsequently, \(\hat{D_{ii}}=\sum_j \hat{A}_{ij}\). The first-order approximation allows the convolution operation to update the representation of each node with its immediate neighbor information, i.e., it is spatially localized. In fact, it can be considered as a bridge between the spectral and spatial methods \citep{zhang2020deep, zhang2019graph}. Therefore, in order to use the information of the k-hop neighbors, several layers of convolution can be stacked.

\subsubsection{Spatial Methods}
While spectral methods are suitable for stationary, simple, and small graphs, spatial methods can be used for dynamic and large graphs with rich feature information. Spatial methods perform filtering operations directly by defining the spatial structures of neighboring nodes in the graph \citep{liu2020introduction, wu2020comprehensive}.

The main challenge of these methods is to define operations for different neighborhood sizes, maintain local stability and weight sharing \citep{liu2020introduction}. To solve these challenges, \cite{NIPS2015_f9be311e} have provided the Neural FPS method, which learns different weight matrices for nodes with different degrees. \cite{niepert2016learning} have extracted a fixed number of neighboring nodes. \cite{hamilton2017inductive} have proposed the GraphSAGE method, which samples a constant-sized neighborhood for each node, then applies an aggregation function (mean or maximal properties of the sample neighbors) to the samples. However, these methods cannot fully utilize the capacity of all neighbors due to sampling from the neighbors. 

\subsection{Graph Structure Learning }
The purpose of GSL is to learn the best representation of the observed data in the form of a graph \citep{subbareddy2019survey,pu2021kernel}. In other words, suppose \(X \in R^{M \times N}\) is the feature matrix, where M is the number of samples and N indicates the number of features, and there is a prior knowledge of data such as data distribution. In this case, relationships between samples can be represented in the form of a graph \citep{dong2019learning}. 

So far, various graph structure learning methods have been developed. \cite{dong2019learning} have been divided into three general approaches, including \textit{statistical or probability-based methods}, \textit{physically-based methods}, and \textit{signal processing-based methods}. 
\begin{itemize}
\item \textit { \textbf{Statistical or probability-based methods}}: The observed data is obtained from a probability distribution. So, the probability distribution is used to model the relationships between the data. This means that there is a graph whose structure is based on the probability distribution of the observed data, such as LDS method \citep{franceschi2019learning}.
\item  \textit {\textbf{Physically-based methods}}: The observed data is the results of some physical phenomena. Thus, the aim is to infer the intrinsic structure of the graph from the physics of the observed data. For example, the information diffusion model on social networks. 
\item \textit {\textbf{Signal processing-based methods}}: The observed data is represented based on its behavior in the field of graph spectrum. In these methods, the goal is to learn a graph that includes certain properties of the observed data, such as signal smoothness or sparse graph. In this category,
smoothness-based methods are one of the simplest and most popular methods. (See \citep{dong2019learning, subbareddy2019survey} for studding other methods.) Smoothness-based methods assume that signals (labels) change smoothly and slowly between adjacent nodes. For example, the temperature  is identical in different places in a geographical area. The smoothness of the signals (labels) is usually measured using Dirichlet energy. Dirichlet energy is based on the quadratic form of the Laplacian matrix. It is calculated for each signal \(x\) in the graph using the following equation:
\begin{equation}
\label{eq:eq9}
\begin{aligned}
tr(X^T LX)=\frac{1}{2} \sum_{i=1}^n \sum_{j=1}^n A_{ij} \|x_i - x_j \|_2^2.
\end{aligned}
\end{equation}
Thus, in these methods, the goal is to find the matrix that minimizes the signal (label) variation on this matrix as follows:
\begin{equation}
\label{eq:eq10}
\begin{aligned}
loss_{gl}= \argmin_{L>0} {tr(X^T L X)+ \lambda f(L)},
\end{aligned}
\end{equation}
where \(f(L)\) is the regularization term to guarantee the learning of the valid matrix (sparse matrix, low-rank matrix, or symmetric matrix).

\item \textbf {Other}: Some research studies use other methods to learn graph. For example, CoGl \citep{shi2021topology} that uses reconstruction error to learn graphs. In this method, the updated feature representation must be close to the original feature representation.
\end{itemize}

In addition to the mentioned categories, structure learning methods can be categorized into two classes, 1)\textit {task-independent learning methods} and 2)\textit {task-driven learning methods}.

The first class learns the optimal graph from the observed data, then use the learned graph in the downstream task (node, graph, or edge classification), such as \citep{kalofolias2016learn,egilmez2017graph,dong2016learning}. However, these methods only produce graphs based on structural information independently downstream tasks. This can reduce performance in combination with graph neural networks \citep{yang2019topology}. In addition, these approaches may create subgraphs with suboptimal structural features because they do not obtain any feedback from the downstream task. The second class follows this key idea that learning a better graph structure depends on better classification of the node (graph or edge), and vice versa. Thus, to achieve the optimal solution, they simultaneously learn the graph structure and the node (graph or edge) classification such as \citep{yu2020graph, jiang2019semi, jin2020graph, yang2019topology,franceschi2019learning, li2018adaptive, pilco2019graph}.  

Also, according to another view, structure learning methods can be categorized into two major groups: 1)\textit {Full graph parameterization} and 2) \textit {Similarity-metric learning}. 

The first group of methods uses learnable parameters to model the weight of edges, such as ProGCN \citep{jin2020graph}, and GLNN \citep{gao2020exploring}. These methods are flexibility designed, but have high memory overhead and time cost, and are not scalable for large graphs due to the over-parameterized problem \citep{yu2020graph, franceschi2019learning,li2018adaptive}. To solve these problems, the second group learns a similarity metric between the pairs of  nodes embedding to model the weight of edges \citep{zhu2021deep}. This increases the learning speed and reduces the number of learnable parameters. Some examples of this group are the GAT \citep{velickovic2018graph} and GLCN\citep{jiang2019semi} methods, which use the attention-based similarity measure to learn the weight of the edges, or the RGLN \citep{tang2021rgln} and AGCN \citep{li2018adaptive} methods, which use the Mahalanobis distance measure.

\subsection{Multi-View Learning}
Multi-view learning is a rapidly growing research field in machine learning. It integrates and learns from multiple views, including multiple relation or various types of features, to improve the generalization performance \citep{wan2022self, xu2013survey, yan2021deep}. 

Views can be integrated in three levels, including \textit {feature}, \textit {relation (structure)} and \textit{results}. In recent years, researchers have proposed many different algorithms based on feature and relation levels. For example, co-training leaning, multiple kernels learning, and subspace learning are among the first approaches for multi-view learning at feature and relation levels \citep{xu2013survey, yan2021deep}. Co-training approaches execute an iterative process to maximize mutual agreement between learners on different views. Multi-kernel approaches identify different kernels corresponding to different views and combine them linearly or nonlinearly. Subspace approaches identify common latent subspaces between different views. \cite{dong2013clustering} and \cite{khan2019multi} believe that subspace learning approaches are better than the other two approaches due to feasibility and satisfactory performance. Therfore, we focus on this approach in this paper .

In addition to these approaches, the emergence of deep learning and graph-based neural networks has led to the formation of another class based on these networks called multi-view GNN \citep{yan2021deep}, such as MR-GCN \citep{huang2020mr}, Relational Graph Convolutional Network (RGCN) \citep{schlichtkrull2018modeling}, and mGCN \citep{ma2019multi}. 

Finally, ensemble learning approaches are among the most popular methods for combining results. Ensemble approaches create a model on each view and then combine their results using different methods such as voting, averaging, and meta learning \citep{sagi2018ensemble}. For example, \cite{keramatfar2022modeling} have proposed a stacking model based on multiple GCNs to leverage the knowledge embedded in different graphs. However, these methods ignore the correlation between views \citep{huang2020mr}.

\section {Related Work} \label{Relatedworks}
In this section, we review recent related works about graph structure learning and multi-view learning based on GCN.
\subsection{Graph Structure Learning }
Graph Attention Network (GAT) is one of the first methods that simultaneously learn graph and prediction task \citep{velickovic2018graph}. GAT does not explicitly generate graphs, but only learns the weight of the relation between each node and its neighbors using the attention-based similarity measure. Therefore, this method cannot add a new edge, but only reweights the edges with a local view. 

Inspired by the GAT, \cite{jiang2019semi} have proposed the GLCN method based on weight calculation between each node and all graph nodes. In fact, this method utilizes a global view of the graph to learn the its general structure. In addition, they have added graph learning loss to the classification loss to ensure that the matrix is valid (sparse and smooth). However, this method also reweights existing edges and cannot add edges.

To solve the edge addition challenge, \cite{yu2020graph} have developed the GRCN method, which includes a revision module for predicting missing edges and reweighting existing edges. Like the previous two methods, this approach uses similarity between pairs of nodes to identify relationships between nodes. However, there is one difference, GRCN utilizes the updated representation of nodes and the dot product function to calculate the similarity. Then, The obtained similarity matrix is added to the original adjacency matrix to insert, delete, and reweight edges. They also select only edges with high confidence to reduce computational costs and prevent excessive noise. However, this method can only improve the existing graph and cannot be used for cases where the graph does not exist.

\cite{chen2020iterative} have transformed the graph structure learning problem into an iterative process of similarity metric learning. They used the multi-head version of the weighted cosine similarity function to calculate the similarity between pairs of nodes and construct the graph. Also, to reduce computational costs and memory consumption, they have developed an anchor-based method that learns the node-anchor affinity matrix instead of learning the similarity matrix between all pairs of nodes.

Similar to previous methods \cite{tang2021rgln} have provided similarities metric learning-based method under a low-rank assumption. They used the Mahalanobis distance measure to identify relationships between nodes. Then, it is optimized by minimizing a similarity-preserving loss. Also, this method considers a low-rank model between features to decompose the similarity matrix into a low-dimensional matrix for implementing and reducing computational costs effectively.

\cite{jin2020graph} have developed a robust graph neural network method based on three characteristics of real graphs, including feature smoothness, low-rank, and sparsity graph to clear the perturbed graph against adversarial attacks.

Like the previous method, \cite{gao2020exploring} have suggested a method to learn the parametrized graph based on the properties of real graphs,including feature smoothness, sparsity graph, loopless and symmetric graph.

\cite{zhan2020graph} mention that point-to-point relationships can not show complex relations. So, they use hypergraph with complex multivariate relations to establish the initial graph. Then, they learn a new parameterized graph based on smoothness of labels and closeness of the new graph to the initial graph.

\cite{lin2021deep} have combined GAT and GLCN methods to present a new method called DGL. 
In this method, the similarity graph is learned from the feature matrix based on the GLCN similarity measure. Then, the learned graph and the feature matrix are fed to the GAT method to update the feature matrix. After applying the GLCN and GAT layers several times, three local and global representations are created for each node. Finally, the weighted sum of the features is used to classify nodes.

\cite{shi2021topology} believe that existing methods seek to adapt the content network and original network structure, while node content and network structure are two distinct but highly correlated data sources. Therefore, they are different in terms of representation learning. As a result, instead of creating compatibility between the two networks, they have proposed a new co-alignment method that models the incompatibility of the two networks for optimal node representation. This method uses a co-training manner so that the content network learns a good node representation for best network reconstruction, and the origin network structure learns an optimal node representation. Like the GAT and GLCN methods, they have used the attention-based similarity measure to calculate the similarity between pairs of nodes.

\cite{peng2021robust} have used the Gaussian kernel to calculate the weight of the edges. They have combined the learned graph with the original graph to use the information in the initial graph to solve the add edge problem of GLCN.

\cite{yang2019topology} believe that adjacent nodes tend to share the same label, so they have modeled the graph structure learning problem as a label propagation process (label smoothness). They have used the dot product of predicted labels to calculate the similarity between pairs of nodes and learn graph structure. 

\cite{li2018adaptive} have proposed the AGCN method, which is the first study for learning graph for each sample. This method learns a residual Laplacian matrix and add it to the original Laplacian matrix. This residual matrix is obtained from the learned adjacency matrix by calculating the Gaussian normalization of the Mahalanobis distance between pairs of nodes.

Unlike these methods, which can only learn one type of relationship between nodes, our proposed method seeks to learn several types of relationships between nodes. Table~\ref{tab:graphlearnig} summarizes the reviewed papers.


\begin{table}[h]
\begin{center}
\begin{minipage}{\textwidth}
\caption{Summery of graph structure learning methods.} \label{tab:graphlearnig}
\begin{tabular*}{\textwidth}{{@{}  p{2cm}  p{2cm}  p{2cm}  p{2cm}  p{2cm}  @{}}}
\toprule%
\textbf{Method} & \textbf{Learning Method} & \textbf{Similarity-Based or Full Graph} & \textbf{Similarity Measure} & \textbf{Task}\\
\midrule
GAT \citep{velickovic2018graph} & - & Similarity-based & Attention-based & Node classification \\
GLCN \citep{jiang2019semi} & Smoothness-based &  Similarity-based & Attention-based & Node classification\\
GRCN \citep{yu2020graph} & Smoothness-based &  Similarity-based & Dot product & Node classification\\
IDGL \citep{chen2020iterative} & Smoothness-based &  Similarity-based & Multi-head weighted cosine similarity & Node and graph  classification\\
RGLN \citep{tang2021rgln} & Smoothness-based &  Similarity-based & Mahalanobis distance & Node classification\\
ProGCN \citep{jin2020graph} & Smoothness-based &  Full graph &  - & Node classification\\
GLNN \citep{gao2020exploring}  & Smoothness-based & Full graph & - & Node classification \\
\citep{zhan2020graph} & Smoothness-based & Full graph & - & Node classification \\
DGL \citep{lin2021deep} & Smoothness-based &  Similarity-based & Attention-based & Node classification\\
CoGL \citep{shi2021topology} & Reconstruction error &  Similarity-based & Attention-based & Node classification\\
DGCN \citep{peng2021robust} & Smoothness-based &  Similarity-based & Gaussian kernel & Node classification\\
To-GCN \citep{yang2019topology} & Smoothness-based & Similarity-based & Dot Product & Node classification\\
AGCN \citep{li2018adaptive} & - &  Similarity-based & Mahalanobis distance & Graph classification\\
\botrule

\end{tabular*}
\end{minipage}
\end{center}
\end{table}

\subsection{Multi-View Learning}
\cite{zhuang2018dual} proposed DualGCN method. This method uses adjacent matrix and positive pointwise mutual information (PPMI) matrix to embed local consistency-based knowledge and global-consistency-based knowledge, respectively. They use a new regularizer function to control the different convolutional results for better label prediction. Note that in this method, the PPMI matrix only helps to better represent the features of adjacent nodes in the adjacency matrix, and no combinations are made at the feature level or graph or results. Also, unlike our method, the graphs are fixed in DualGCN.

\cite{schlichtkrull2018modeling} have proposed the RGCN method for multi-relational graphs by integrating features. In this method, after applying a GCN on each relation and updating the features, the average of the updated features on each node is used to predict the label of nodes.

\cite{ma2019multi} have suggested the mGCN method at the feature level. They believe that each node has two types of relationships. First, the relationship between each node and its neighbors in each dimension (within-dimension interactions). Second, the relationship between each node and its copy in the other dimensions (across-dimension interactions).  Thus, they create a specific-dimension representation for each node by combining the within- and across-dimension representations. Finally, they concatenate the specific-dimension representations to create a general representation for each node. This representation can be used to predict the label of nodes. 

\cite{khan2019multi} have proposed a Multi-GCN method for multi-view learning. They have used Grassmann learning to merge different graphs. However, the used graphs in this method are static.

\cite{lin2020structure} have proposed the SF-GCN method, which fusions different graphs by exploring the common and specific properties of structures. Unlike the previous method, SF-GCN uses the weighted sum of the graphs and consideres the importance of each structure in the fusion process. It uses Grassmann learning for obtaining the weight of the importance of each graph.

\cite{wang2020gcn} argue that the GCN method cannot adaptively integrate the information contained in the feature space and structure space. Therefore, they have proposed a multi-channel method called AM-GCN for combining this information. In addition to the topology graph, this method creates a feature graph using the cosine similarity based on the features of the nodes. Then, two GCN networks are created on the feature graph and the topology graph, and a shared GCN network is created using two graphs. The importance of the features obtained from these networks is identified using the attention mechanism. Finally, the extracted features are combined based on the attention weight for the classification task. This method also uses static graphs, unlike our proposed method.

\cite{huang2020mr} have proposed a special convolutional operation for multi-relation graphs based on the eigen-decomposition of a Laplacian tensor, which takes into account the correlations across the relations. The eigen-decomposition is formulated with a generalized tensor product, which can correspond to any unitary transform instead of limited merely to Fourier transform. This method also used fixed graphs during the training.

\cite{yu2020graph} theoretically show that the GRCN method can also be used for multiple graphs, due to the use of the addition operator. Therefore, an integrated matrix can be created according to the following steps. Firstly, GCN is applied on each graph; then, similarity matrices are calculated using the dot product of the new representation related to each node; finally, similarity matrices and initial matrices are summed together. This method also uses only one similarity measure to create multiple dynamic graphs. In addition, it uses only the classification loss function, which may not guarantee a valid graph.

\cite{peng2021robust} have presented the DGCN method, which combines both the features and the graph. Firstly, DGCN calculates the similarity graph of each view using the Gaussian kernel. Then, in order to benefit from the comprehensive information of all views, the weighted average of the learned graph corresponding to each view and the initial graphs of all views are fed into \(M\) GCN networks. Finally, the updated representations of each node are combined for node classification. Although this method uses the dynamic graph, unlike our method, only one graph is learned from a view, and the learned graph is combined with the existing original graphs.

\cite{adaloglou2020multi} have presented the MV-AGC method for multi-relational graphs. MV-AGC is an extended version of AGCN \citep{li2018adaptive}. In this method, first a few similarity graphs are learned using the Mahalanobis distance measure. Then, in order to maintain the original structure of the graph, the initial Laplacian matrix is added to the learned Laplacian matrix. Then, multiple GCNs are used to create multiple new representations for each node. Finally, the maximum of normalized representations of each node is used as the final representation to predict the label. Similar to our method, this method learns different relationships between pairs of nodes. However, it differs from our proposed method in several aspects. First, it is used to classify graphs, while we focus on node classification. Second, it utilizes a single learning method (Mahalanobis similarity measure) to create multiple graphs, which may not guarantee the different types of relationships. Third, since it just uses the classification loss function, it may not guarantee valid graph learning; while we include sparseness and smoothness of the learned graphs in the proposed method. 
Table~\ref{tab:viewlearning} summarizes the reviewed articles.

\begin{table}[h]
\begin{center}
\begin{minipage}{\textwidth}
\caption{Summery of multi-view learning methods.}\label{tab:viewlearning}
\begin{tabular*}{\textwidth}{{@{} p{1.8cm} p{1cm} p{1.8cm} p{1.8cm}  p{1.9cm}  p{1.7cm}  @{}}}
\toprule%
\textbf{Method} & \textbf{Graph Type} & \textbf{Aggregation Method}  & \textbf{Aggregation Level} & \textbf{Weighted or Unweighted Aggregation} & \textbf{Weighted Method}\\
\midrule
DualGCN \citep{zhuang2018dual} & Static & - &- & - & -\\
RGCN \citep{schlichtkrull2018modeling} & Static  &  Average & Feature & Unweighted & -\\
mGCN \citep{ma2019multi} & Static & Concatenate &  Feature & Weighted & Fully connected layer \\
Multi-GCN \citep{khan2019multi} & Static &  Grassmann learning & Structure  & Unweighted & -\\
SF-GCN \citep{lin2020structure} & Static &  Sum and Multiple & Structure & Weighted & Grassmann learning\\
AM-GCN \citep{wang2020gcn} & Static &  Sum & Feature & Weighted & Attention\\
MR-GCN \citep{huang2020mr} & Static & Multi-relational convolution operator & Feature & Unweighted & -\\ 
GRCN \citep{yu2020graph} & Dynamic & Sum  & Structure & Unweighted & - \\
DGCN \citep{peng2021robust}  & Dynamic &  Average and Sum & Structure and Feature & Weighted & Attention\\
MV-AGC \citep{adaloglou2020multi} & Dynamic	& Sum and Max &	Structure and Feature &	Unweighted & -\\
\botrule
\end{tabular*}
\end{minipage}
\end{center}
\end{table}

\section{Proposed Method} \label{ProposedFramework}
While learning multiple relationships between nodes can provide more comprehensive knowledge of the data and improve model performance, most existing methods learn a single relationship between pairs of nodes.
In this section, we present the details of our multi-view graph structure learning approach, MV-GSL, which exploits several kinds of relationships between pairs of nodes. 

\begin{figure}[ht]%
\centering
\includegraphics[width=1\textwidth]{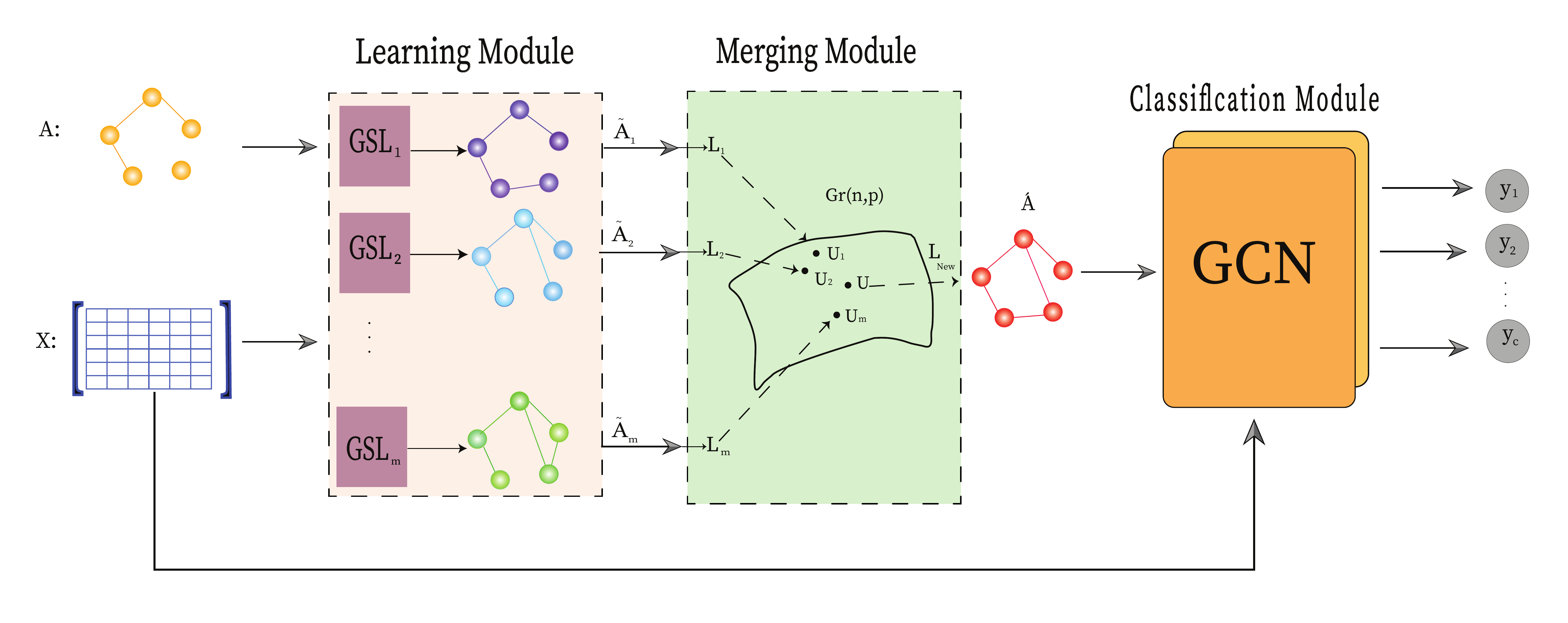}
\caption{\label{fig:framework} Framework of the proposed MV-GSL method.}
\end{figure}

As shown in Figure \ref {fig:framework}, the proposed framework consists of three modules: \textit {learning}, \textit {merging}, and \textit {classification}.  
The learning module utilizes some of the single-graph structure learning methods to obtain multiple graph structures. The merging module unifies the learned graphs by subspace merging using Grassmann manifold. Finally, the classification module utilizes the unified graph to classify nodes. The details of each module are described in the following.

\subsection{Learning }
The purpose of this module is to create multiple graphs to be used as multiple views. 
For this purpose, 
different single-graph structure learning methods 
are used to learn $m$ different graphs $\tilde {A}_i, 1 \leq i \leq m$. 
Each method can use two knowledge sources, the feature matrix \(X\) and the graph topology matrix \(A\). 
For effective merging, the base graphs should be diverse and complementary. 
In this paper, we exploit five methods: GAT \citep{velickovic2018graph}, GLCN\citep{jiang2019semi}, RGLN \citep{tang2021rgln}, GRCN \citep{yu2020graph}, and ProGCN \citep{jin2020graph}. 

The GAT method learns the graph structure with attention-based similarity metric with a local view as

\begin{equation}
\label{eq:eq11}
\begin{aligned}
\tilde{A}_{ij}=\frac{\exp( \textrm{leakyReLU}(a^T (Wx_i \| Wx_j)))}{\sum_{k \in N_i}{\exp(\textrm{leakyReLU}(a^T (Wx_i \| Wx_k)))}},
\end{aligned}
\end{equation}
where \( \|\) is the concatenation operator, \(W\) is the parameter matrix for the shared linear transform, $a$ is the weighted vector of the shared attention, \(N_i\) is the set of neighbors of node $i$, and LeakyRelu is the nonlinear activation function.

The GCLN method learns the graph structure with attention-based similarity metric similar to GAT but with a global view as 
\begin{equation}
\label{eq:eq12}
\begin{aligned}
\tilde{A}_{ij}=\frac{A_{ij} \exp(\textrm{ReLU}(a^T (Wx_i - Wx_j)))}{\sum_{k=1}^n {A_{ik}\exp(\textrm{ReLU}(a^T (Wx_i - Wx_k)))}}, 
\end{aligned}
\end{equation}
where \(A_{ij}\) is initial graph. It also minimizes the following loss function along with the classification loss to ensure learning of a valid graph:

\begin{equation}
\label{eq:eq13}
\begin{aligned}
loss_{gl}=\sum_{i=1}^n \sum_{j=1}^n A_{ij} \|x_i - x_j \|_2^2 + \alpha \|\tilde{A}\|_F^2,
\end{aligned}
\end{equation}
where the first term ensures the smoothness of adjacent signals and the second term controls the sparsity of the learned graph.

The RGLN method learns the graph structure with Mahalanobis distance metric as
\begin{equation}
\label{eq:eq14}
\begin{aligned}
\tilde{A}_{ij}= A_{ij}+ \exp(-\|R^T (x_i - x_j)\|_2^2),
\end{aligned}
\end{equation}
where \(R \in R^{d \times s}\) is a low-rank weight matrix, \( s\ll d\) which significantly reduces the number of the learnable parameters due to low-rank property, and \(A_{ij}\) is initial graph. It also ensures learning of a valid graph by minimizing the smoothness constraint as
\begin{equation}
\label{eq:eq15}
\begin{aligned}
loss_{gl}=\frac{1}{2} \sum_{i=1}^n \sum_{j=1}^n A_{ij} \|x_i - x_j \|_2^2.
\end{aligned}
\end{equation}

The GRCN method modifies the original graph with the dot product of the updated node representations after using GCN as follow:

\begin{equation}
\label{eq:eq16}
\begin{aligned}
Z=\textrm{GCN}(A,X),   \\
\tilde{A}=A+ \textrm{dot}(Z,Z).
\end{aligned}
\end{equation}

The ProGCN method learns a full parametric matrix by optimizing the following loss function:
\begin{equation}
\label{eq:eq17}
\begin{aligned}
loss_{gl}=\sum_{i=1}^n \sum_{j=1}^n A_{ij} \|x_i - x_j \|_2^2 +\alpha \|\tilde{A}\|_1 + \beta \|\tilde{A}\|_* +\gamma \|A-\tilde{A}\|_2^2,
\end{aligned}
\end{equation}
where the terms control smoothness, sparsity, low-rankness, and closeness of the learned graph to the original graph, respectively.
Also \(\alpha\), \(\beta\), and \(\gamma\) are hyperparameters that determine the contributions of the constraints.

\subsection{Merging}\label{subsec:merging}
The output of the learning module is \( m\) matrices \( \{\tilde {A}_1, \tilde {A} _2, … \tilde {A}_m \} \), where each matrix \( \tilde {A}_i \in R^{n \times n}\) is a learned graph. We use the Grassmann learning to merge the learned graphs to obtain an informative combination of the base graphs~\citep{dong2013clustering}.
The usage of Grassmann learning retains the specific structural properties of each learned graph and creates a common structure between graphs rather than mixing them.

The Grassmann manifold is a special manifold related to Euclidean space  \citep{bendokat2020grassmann,dong2013clustering}. Mathematically, the Grassmann manifold \( Gr(n,p)\) is the space of $n$-by-$p$ matrices (e.g., \(Y\)) with orthonormal columns, where \(0 \leq p \leq n \) , i.e.,
\begin{equation}
\label{eq:eq18}
\begin{aligned}
Gr(n,p)= \{Y \mid Y \in R^{n \times p },Y^TY=I \}.
\end{aligned}
\end{equation}

According to Grassmann theory, each orthonormal matrix forms a unique subspace, so it can be mapped to a unique point in the Grassmann manifold \citep{lin2020structure}. Since the eigenvector matrix of the normalized Laplacian matrix (\(U \in R^{n \times p}\)), which contains the first \( p\) eigenvectors, 
is orthonormal \citep{wu2020comprehensive}, it also forms a unique subspace that can map  a single point on the Grassmann manifold. 

On the other hand, each row of the eigenvector matrix represents the spectral embedding of each node in the \(p\)-dimensional space. So, the two adjacent nodes have close embedding vectors. This subspace representation, which summarizes graph information, can be used for a variety of tasks, such as clustering, classification, and merging graphs \citep{dong2013clustering}.

We use subspace representation for merging graphs. 
The integrated subspace \(U\) 
should have the shortest distance to all subspaces $\{U_i\}_{i=1}^m$, 
while preserving the connections between the nodes in each individual subspace as much as possible. This is obtained using the following objective function: 
\begin{equation}
\label{eq:eq19}
\begin{aligned}
\min_{U \in R^{n \times p}} \sum_{i=1}^m \textrm {tr}(U^T L_i U) + \alpha (p \times m - \sum_{i=1}^m \textrm {tr} (U U^T U_i U_i^T)), \quad s.t. \quad U^T U=I,
\end{aligned}
\end{equation}

where \(m\) is the number of graphs and \(\alpha\) is a hyperparameter to control the relative importance of two losses.
In this equation, the first term controls the node connectivity based on spectral embedding, and the second term controls the distance between the merged subspace and the individual subspaces. The second term is based on the projection distance between the principal angles of subspaces \( \theta _{ij} \) as follows:

\begin{equation}
\label{eq:eq20}
\begin{aligned}
\sum_{i=1}^m d^2 (U,U_i)&= \sum_{i=1}^m \sum_{j=1}^p \sin^2 \theta_{ij} \\
&=\sum_{i=1}^m (p- \sum_{j=1}^p \cos^2 \theta_{ij})\\ 
&=\sum_{i=1}^m (p-\textrm{tr}(UU^T U_i U_i^T ))\\
&=p\times m- \sum_{i=1}^m \textrm{tr}(UU^T U_i U_i^T ).
\end{aligned} 
\end{equation}

Ignoring the constant value \(p\times m\) and solving this optimization problem using the Rayleigh-Ritz theorem, the new Laplacian matrix is obtained as 
\begin{equation}
\label{eq:eq21}
\begin{aligned}
L_{new}=\sum_{i=1}^M L_i -\alpha \sum_{i=1}^M U_i U_i^T.
\end{aligned}
\end{equation}

After calculating the eigenvectors matrix \(U_i\) corresponding to the Laplacian matrix \(L_i\) of each graph \( \tilde{A}_i\) and calculating the aggregated Laplacian matrix, the aggregated adjacency matrix is obtained as follows:
\begin{equation}
\label{eq:eq22}
\begin{aligned}
 \Acute{A} =D-L_{new}.
\end{aligned}
\end{equation}

Aggregated graphs may have negative values due to the presence of a correlation between eigenvectors, while real graphs have normally positive edge weights. Therefore, the ReLU function is used to remove the negative weights while keeping the positive weights. 

The integrated graph may be very dense, causing high computational complexity for the classification module and low accuracy due to noisy edges. 
To resolve this issue,  we use KNN sparsification, 
where just the $k$ edges with the highest value are kept for each node. Therefore, the integrated graph is updated as:
\begin{equation}
\label{eq:eq23}
\begin{aligned}
\tilde{A}_{ij}= \begin{cases}
			\Acute{A}_{ij}, & \text{if  $\Acute{A}_{ij} \in Top\_k $ }\\
            0, & \text{otherwise.} 
		 \end{cases} 
\end{aligned}
\end{equation}

Finally, the final graph is made symmetric using the following equation: 
\begin{equation}
\label{eq:eq24}
\begin{aligned}
\hat{A} =\frac{(\tilde{A} +\tilde{A}^T )}{2}.
\end{aligned}
\end{equation}

\subsection{Classification}
For the classification module, we adopt a two-layered GCN \citep{kipf2017semi} as follows: 
\begin{equation}
\label{eq:eq25}
\begin{aligned}
Z=\textrm{Softmax}(\Bar{A} \textrm{ReLU} (\Bar{A}XW^0 ) W^1),
\end{aligned}
\end{equation}
where \( \Bar{A}=I_n+\hat{D}^{\frac{-1} {2}} \hat{A} \hat{D}^{ \frac{-1}{2}}\), and \(W^0\) and \(W^1\) are weight matrices. Softmax and ReLU are activation functions in the hidden and output layers, respectively. In this equation, the first layer creates a new representation of the original raw features \(X\), while the second layer turns the created intermediate representation into the final representation $Z$. The optimal weight matrices are obtained by minimizing the cross-entropy loss function as follows:
\begin{equation}
\label{eq:eq26}
\begin{aligned}
loss_{CE}=\sum_{i=1}^n \sum_{j=1}^c y_{ij}\ln z_{ij}.
\end{aligned}
\end{equation}

\section{Experiments } \label{Experiments}
We conduct extensive experiments to 
compare the proposed method with previous state-of-the-art methods in terms of classification accuracy. For base methods, the results are obtained using publicly available codes. All the reported results are averaged over five runs. 
We compare the proposed MV-GSL method with both single-graph and multi-graph structure learning methods. 
We also investigate different methods based on their matrix representations. 

\subsection {Setting}
The node classification task is used to evaluate performances. 
In this regard, we use two popular paper citation network datasets, Cora~\citep{sen2008collective} and Citeseer \citep{sen2008collective}\footnote{The datasets are available at \url{https://github.com/kimiyoung/planetoid/tree/master/data}}. In these datasets, the nodes represent the papers that were published in a journal, and the edges show citations. Each publication is described by a sparse one-hot feature vector, which indicates the absence or presence of the corresponding word from a learned dictionary. Table~\ref{tab:dataset} summarizes the statistics of the citation datasets. As shown in the table, 
both have low label rates making them appropriate semi-supervised datasets which are commonly used. We use a transductive setting, which assumes all unlabeled data are available at the training time. 

\begin{table}[ht]
\begin{center}
\begin{minipage}{\textwidth} 
\caption{The statistics of datasets used in the experiments.}\label{tab:dataset}
\begin{tabular*}{\textwidth}{{@{}p{1cm} p{1.1cm} p{1cm} p{1cm} p{1cm} p{1cm} p{1cm} p{0.7cm} p{0.9cm} @{}}}
\toprule%
\textbf{Dataset} & \textbf{ \verb|\#|Node} & \textbf{ \verb|\#|Edge} & \textbf{\verb|\#|Feature} & \textbf{\verb|\#|Class} & \textbf{Label rate} & \textbf{\verb|\#|Train} & \textbf{\verb|\#|Val} & \textbf{\verb|\#|Test} \\
\midrule
Cora & 2708 & 5429 & 1433 & 7 & 0.052 & 140 & 500 & 1000\\
Citeseer & 3327 & 4732 & 3703 & 6 & 0.0036 & 120 & 500 & 1000 \\
\botrule
\end{tabular*}
\end{minipage}
\end{center}
\end{table}

We work on five popular state-of-the-art single-graph structure learning method: GLCN \citep{jiang2019semi}, GAT \citep{velickovic2018graph}, RGLN \citep{tang2021rgln}, GRCN \citep{yu2020graph}, and ProGCN \citep{jin2020graph}. 
To satisfy the requirements of input matrices in the Grassmann method, we make the matrices symmetric and remove non-zero diagonal elements. 

In the experiments, we use the train-test split used in 
\cite{kipf2017semi}. For the methods that already used this split, including GAT\footnote{\url{https://github.com/PetarV-/GAT.git}}, RGLN\footnote{\url{https://github.com/ashawkey/JLGCN.git}}, and GRCN\footnote{\url{https://github.com/PlusRoss/GRCN.git}}, we use the same hyperparameter values presented in their papers. 
For methods that used a different train-test split, including GLCN\footnote{\url{https://github.com/jiangboahu/GLCN-tf.git}} and ProGCN\footnote{\url{https://github.com/ChandlerBang/Pro-GNN.git}}, we optimize the hyperparameter values to make a fair comparison. 

We use the random search method on the validation dataset for tuning hyperparameters. In the merging module, the size of subspace is set to 10 times number of classes and the regularization parameter \(\alpha\) is chosen from $\{0.1, 0.2, \ldots, 0.9\}$. 
In the classification module, we choose the learning rate from 
$\{0.01, 0.03, 0.1, 0.2, 0.3, 0.5\}$, number of hidden units from $\{16, 32, 64, 128, 256, 512\}$, dropout rate from $\{0.5, 0.6, 0.7, 0.8, 0.9\}$, and weight decay from $\{5e-3, 5e-4, 5e-5\}$. The number of epochs is 5000, and the learning process stops when the training loss is higher than the average loss in the last 10 epochs. 

\subsection {Results and Discussion}

Table~\ref{tab:result} shows the results of comparing the proposed MV-GSL method with the five sing-graph learning methods as well as four multi-graph methods. The used multi-graph methods are
\begin{itemize}
\item \textit{Average} which uses the average of the symmetric normalized adjacency matrices \( (\frac{1}{m} \sum_{i=1}^m \tilde{D}^\frac{-1}{2} \tilde{A} \tilde{D}^\frac{-1}{2}) \),
\item \textit {Ensemble} which aggregates the outputs of different GCN models, using either average or maximum probability, instead of merging the graphs,
\item \textit{RGCN} \citep{schlichtkrull2018modeling} which uses the average of features. 
\end{itemize}

As shown in this table, the MV-GSL method outperforms all single and multi-graph methods on both datasets. 

\begin{table}[!htbp]
\begin{center}
\begin{minipage}{0.7\textwidth}
\caption{The accuracy of different single and multi-graph methods on two datasets Cora and Citeseer. Best results are in \textbf{bold}.}\label{tab:result}
\begin{tabular*}{\textwidth}{{@{}lll@{}}}
\toprule%
\textbf{Method} & \textbf{Cora} & \textbf{Citeseer} \\
\midrule
GAT \citep{velickovic2018graph} & \(83.5\pm 0.5 \) & \(73.0\pm 0.5 \)  \\
GLCN \citep{jiang2019semi} & \(82.0\pm 0.3 \) & \(71.8\pm 1.3 \)  \\
GRCN \citep{yu2020graph} & \(84.1\pm 0.3 \) & \(73.0 \pm 0.5 \) \\
RGLN \citep{tang2021rgln} & \(81.5\pm 0.7 \) & \(73.5 \pm 1.7 \)  \\
ProGCN \citep{jin2020graph} & \(81.0\pm 0.3 \) & \(68.4 \pm 0.6 \)  \\
\midrule
Average & \(83.2\pm 0.3 \) & \(73.4\pm 0.2 \)  \\
Ensemble (Average Prob)  & \(83.5\pm 0.3 \) & \(71.8\pm 0.4 \)  \\
Ensemble (Max Prob) & \(83.2\pm 0.7 \) & \(71.1 \pm 1.1 \)  \\
RGCN \citep{schlichtkrull2018modeling} & \(82.7\pm 1.1 \) & \(72.7 \pm 0.6 \)  \\
\midrule
MV-GSL (proposed method)
& \pmb{\(84.9\pm 0.6 \)} & \pmb{\(74.0 \pm 0.4 \)}  \\
\botrule
\end{tabular*}
\end{minipage}
\end{center}
\end{table}

\begin{figure}[!htbp]%
\centering
\includegraphics[width=0.98\textwidth]{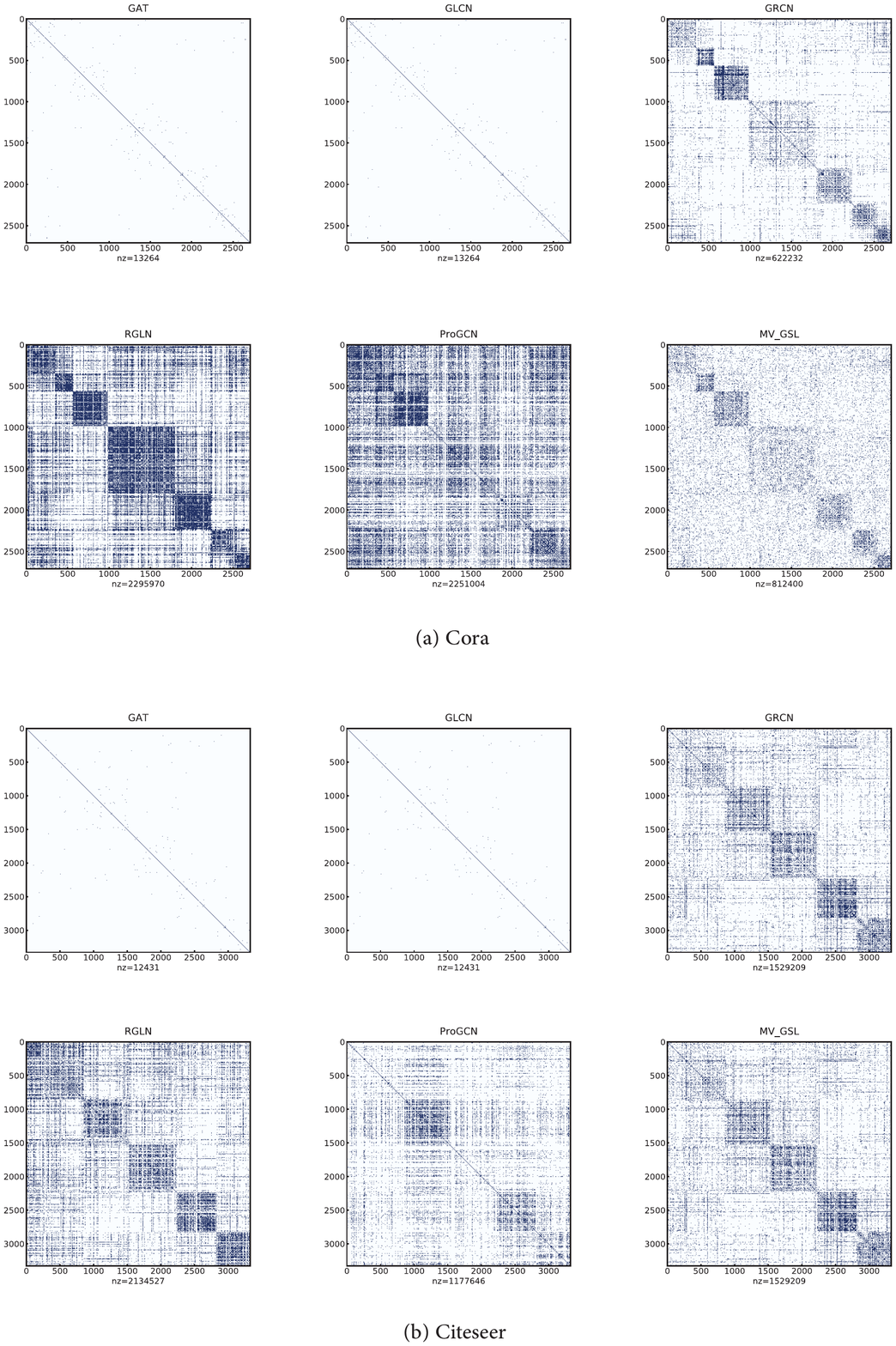}
\caption{Spy plots of the adjacency matrices. RGLN matrix is displayed with the precision of 0.0001.}\label{fig:Matrix}
\end{figure}

To visualize the learned graphs, we show the spy plot of the adjacency matrices in Figures~\ref{fig:Matrix}. 
It visualizes the non-zero values (\(nz\)) of the matrices, ordered such that the samples of each class are next to each other. 
It is clear that, GAT and GLCN just reweight the edges, as previously stated. RGLN create a fully connected matrix, although most edges are light weight. RGLN and GRCN establishing good intra-class relationships. However, they has added a lot of inter-class relationships, which increases noise. 
ProGCN has identified more inter-class relationships in both datasets than other methods, which has caused a reduction in performance. 
On the other hand, our method provides better results than the base methods in both datasets by reducing inter-class relations and reducing false edges.

\begin{figure}[ht]%
\centering
\includegraphics[width=0.9\textwidth]{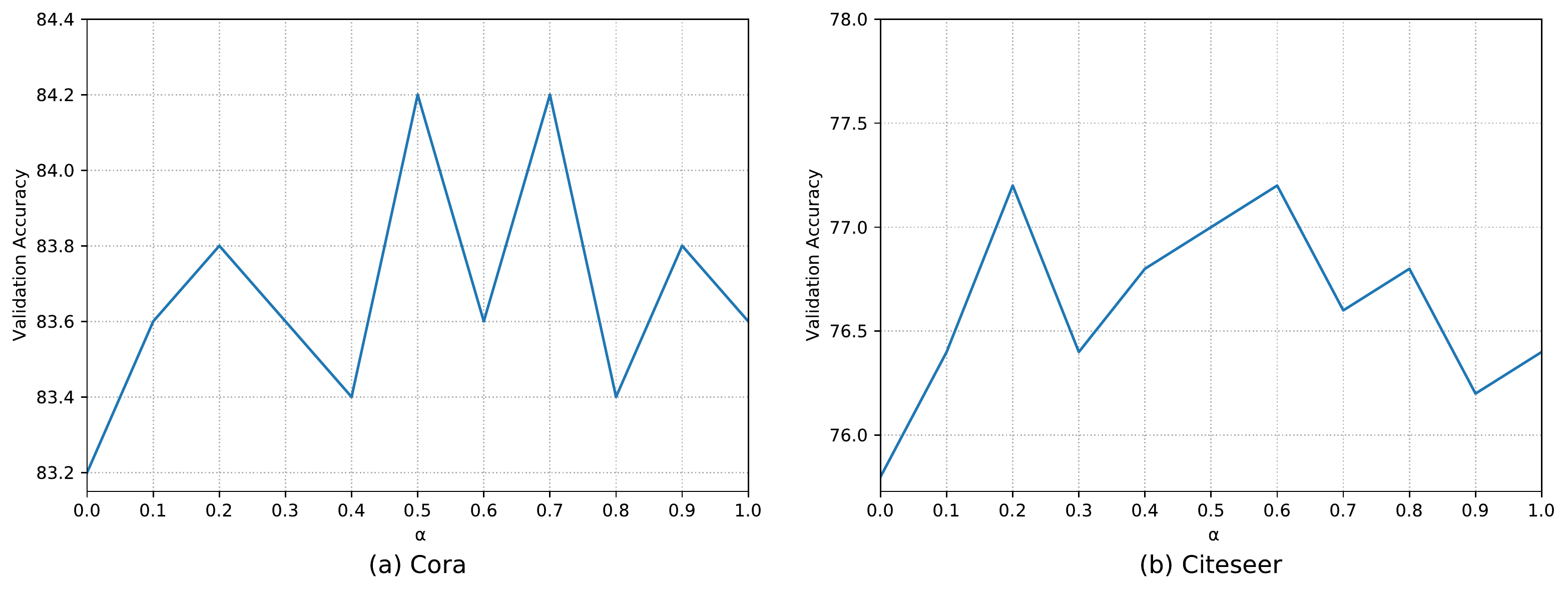}
\caption{
The effect of changing the value of \(\alpha\) on the performance of MV-GSL method.}\label{fig:Alphachange}
\end{figure}

In the next experiment, we investigate the effect of changing the value of trade-off parameter $\alpha$ in Equation~\ref{eq:eq21}. As mentioned in subsection~\ref{subsec:merging}, the merging module combines the sum of Laplacian matrices with a ratio of the new correlation matrix to create the merged matrix. This correlation matrix is created based on spectral embedding of the nodes. When \(\alpha\) is 0, 
the merged matrix is equivalent to the sum of Laplacian matrices. Figure \ref {fig:Alphachange} shows the results of changing this parameter. According to this figure, the parameter \(\alpha\) plays an important role in the classification accuracy, which shows that the correlation matrix has a significant effect on the merged matrix. 

\begin{figure}[h]%
\centering
\includegraphics[width=0.6\textwidth]{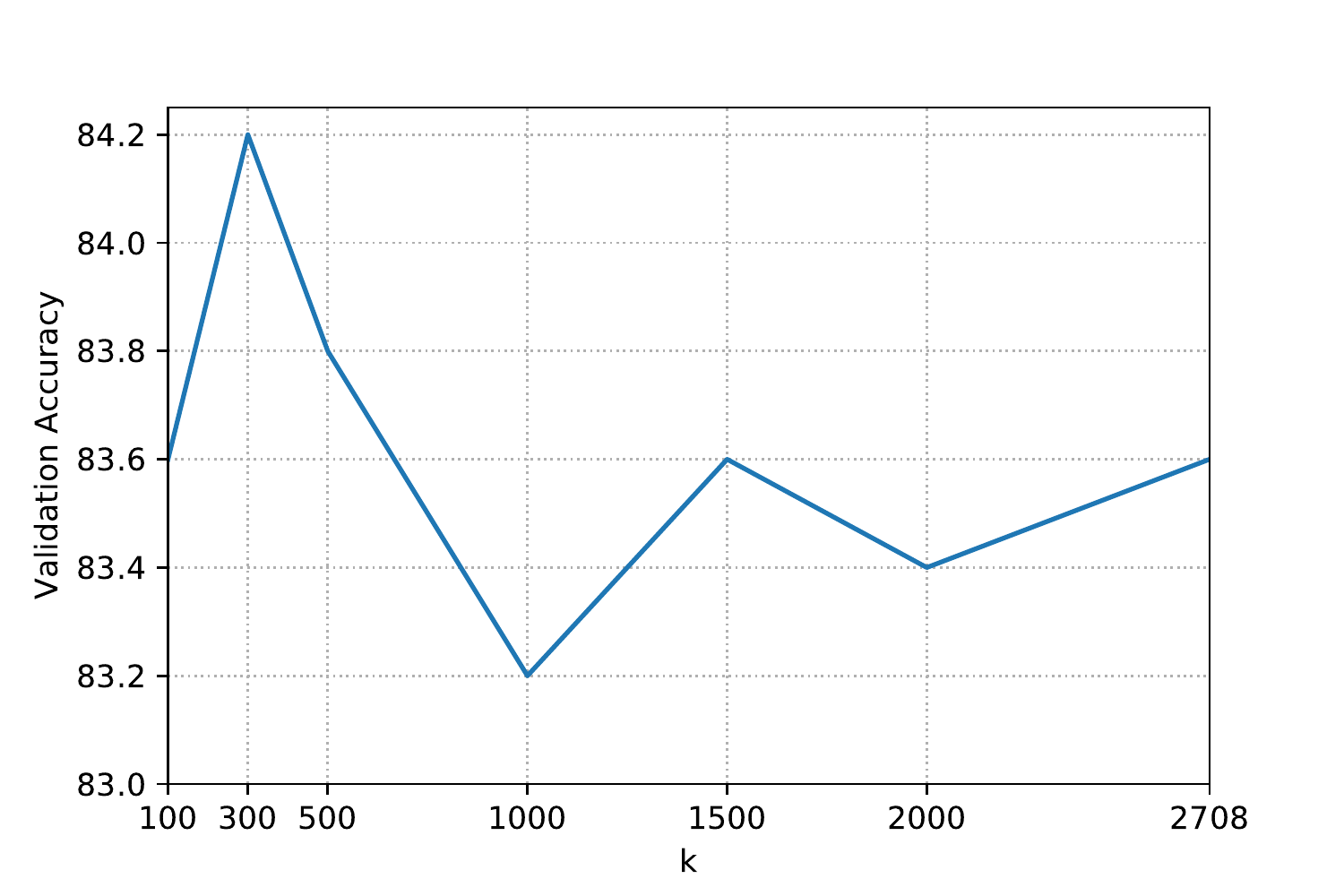}
\caption{\label{fig:Sparseification} 
The effect of changing the value of sparsification parameter \(k\) on the performance of MV-GSL method.}
\end{figure}

Finally, we examine the effect of the sparsification parameter (\(k\)). 
As shown in Figure \ref {fig:Sparseification}, this parameter has a significant effect on the performance of MV-GSL method. 
As mentioned earlier, applying the sparsification not only improves the performance of the model but also drastically reduces the computational cost.

\section{Conclusion} \label{Conclusion}
In this paper, we proposed a multi-relational graph structure learning method. 
The proposed framework learns multiple relationships and performs node classification simultaneously. 
It obtains higher accuracies by creating different relationships between nodes. 
It also can be used for applications that need to learn several kinds of relationships between nodes. 
To verify the effectiveness of the proposed method, extensive experiments were performed on two popular datasets, Cora and Citeseer. The experiments showed that the proposed method 
obtains better results compared to single-relational methods as well as other integrating methods.

The proposed method has also some limitations which can be investigated in future work. First, it is modular which can lead to suboptimal results. 
In other words, it learns each graph 
in isolation and keeps the learned graphs fixed during integration. 
An end-to-end approach which improves the learned graphs 
during merging can obtain better results. Second, 
different relationships between nodes can have different effects on the final result. So, weighting the relationships to consider their importance can lead to further improvements. 

\bibliography{sn-bibliography}


\end{document}